\documentclass[conference]{IEEEtran}
\IEEEoverridecommandlockouts
\usepackage{cite}
\usepackage{amsmath,amssymb,amsfonts}
\usepackage{algorithmic}
\usepackage{graphicx}
\usepackage{textcomp}
\usepackage{xcolor}
\usepackage{wrapfig}
\graphicspath{ {./images/} }

\def\BibTeX{{\rm B\kern-.05em{\sc i\kern-.025em b}\kern-.08em
    T\kern-.1667em\lower.7ex\hbox{E}\kern-.125emX}}
\begin{document}

\title{Maximum Relevance and Minimum Redundancy Feature Selection Methods for a Marketing Machine Learning Platform}

\author{
\IEEEauthorblockN{Zhenyu Zhao}
\IEEEauthorblockA{\textit{Uber Technologies, Inc.} \\
San Francisco, USA \\
zhenyuz@uber.com}
\and
\IEEEauthorblockN{Radhika Anand}
\IEEEauthorblockA{\textit{Uber Technologies, Inc.} \\
San Francisco, USA \\
radhika@uber.com}
\and
\IEEEauthorblockN{Mallory Wang}
\IEEEauthorblockA{\textit{Uber Technologies, Inc.} \\
San Francisco, USA \\
mallory.wang@uber.com}
}

\maketitle

\begin{abstract}
In machine learning applications for online product offerings and marketing strategies, there are often hundreds or thousands of features available to build such models. 
Feature selection is one essential method in such applications for multiple objectives: improving the prediction accuracy by eliminating irrelevant features,  accelerating the model training and prediction speed, reducing the monitoring and maintenance workload for feature data pipeline, and providing better model interpretation and diagnosis capability.
However, selecting an optimal feature subset from a large feature space is considered as an NP-complete problem. The mRMR (Minimum Redundancy and Maximum Relevance) feature selection framework solves this problem by selecting the relevant features while controlling for the redundancy within the selected features.
This paper describes the approach to extend, evaluate, and implement the mRMR feature selection methods for classification problem in a marketing machine learning platform at Uber that automates creation and deployment of targeting and personalization models at scale.
This study first extends the existing mRMR methods by introducing a non-linear feature redundancy measure and a model-based feature relevance measure. Then an extensive empirical evaluation is performed for eight different feature selection methods, using one synthetic dataset and three real-world marketing datasets at Uber to cover different use cases. Based on the empirical results, the selected mRMR method is implemented in production for the marketing machine learning platform. A description of the production implementation is provided and an online experiment deployed through the platform is discussed.
\end{abstract}

\begin{IEEEkeywords}
feature selection, automated machine learning, marketing, classification, random forest
\end{IEEEkeywords}

\section{Introduction}
Nowadays, machine learning is being broadly used in the digital marketing field to communicate the product or services to existing or potential customers through advertisements, in-product recommendations, customer care, and other channels \cite{katsov2017introduction}. In particular, the classification models can be used in a wide range of applications, such as: (1) user acquisition: identifying potential customers by estimating the likelihood of a new user to sign up for a product; (2) cross-sell or up-sell: recommending related products or services to existing users by calculating the propensity of a user to use the certain product or services; (3) user churn prediction and reactivation: predicting the probability of a user to churn (becoming inactive), in order to plan for re-engagement strategy.

In such machine learning applications, obtaining and storing a large set of features has become easy and low-cost in recent years, through logging user online activities in distributed data file system. On one hand, such rich feature set, if used wisely, can provide advantage in model performance. On the other hand, directly using all the available features in the model can lead to computation inefficiency, over-fitting, high maintenance workload, and model interpretation difficulty. 

Feature selection is an essential step in such large-scale machine learning applications to leverage the benefits of rich feature store and overcome the associated challenges and costs. It improves the machine learning application and system in multiple ways: (1) Faster computation speed: with a smaller set of features, the model training and prediction process is accelerated; (2) More accurate prediction: this is achieved by multiple means: eliminating irrelevant features, avoiding over-fitting, and fitting more training sample into memory thanks to reduced number of features. (3) Lower maintenance cost for feature data pipeline: reducing the number of features can significantly reduce the cost for creating, monitoring, and maintaining the model feature pipeline.  
(4) Easier model interpretation and diagnosis: by only including the important feature set in the modeling process, it is easier to interpret what features and information the model prediction is based on. It enables easier check of model compliance for marketing targeting policy, better business intuition and insights, as well as a more focused diagnosis space on a smaller feature set if the model encounters issues. 

Various feature selection methods (\cite{bolon2013review,chandrashekar2014survey,tang2014feature}) are available to reduce the entire feature set to a more compact one. 
Such methods can be roughly put into three categories: filter methods, wrapper methods, and embedded methods.
In this paper, a filter method called mRMR is studied and evaluated to be implemented in an automated machine learning platform. The advantage of the filter method is the computation efficiency and generalizability to different machine learning models. Out of filter methods, the motivation to use mRMR method is that it can effectively reduce the redundant features while keeping the relevant features for the model. 
It is known that “the m best features are not the best m features” \cite{cover1974best}, with the reason that many important features are correlated and redundant. The mRMR method solves this problem by selecting features considering both the relevance for predicting the outcome variable and the redundancy within the selected features.

The mRMR method was initially developed by Peng et al. \cite{peng2005feature} for pattern classification systems, that uses mutual information to measure both the relevance and the redundancy. Several variants of mRMR method were developed afterwards. 
To make a better balance between the relevance measure and redundancy measure by mutual information, a normalized mutual information feature selection (NMIFS) method \cite{estevez2009normalized} is developed  to normalize the redundancy term. Based on NMIFS, a related approach is proposed \cite{thang2010improved} to further normalize the relevancy term. Given the different information based feature selection methods, a conditional likelihood framework was proposed to unify these approaches by concluding: many heuristics for information based feature selection methods can be regarded as iterative maximizers of the conditional
likelihood \cite{brown2012conditional}. To overcome the computation complexity of mRMR, a Fast-mRMR implementation framework in distributed systems is discussed in \cite{ramirez2017fast}. In addition to the information-based measure, other variants of feature selection methods have been developed under the mRMR framework for making trade-offs between relevance and redundancy. 
One variant with better computation efficiency is proposed by \cite{ding2005minimum}: using F-statistic to measure the relevance and correlation to measure the redundancy.
The maximum relevance–minimum multicollinearity (MRmMC) method, proposed by \cite{senawi2017new}, measures the relevance by conditional variance and eliminates
redundancy through multiple correlation with an orthogonal projection scheme. One related study applies mRMR method in an online user modeling use case and evaluates the mRMR performance with different machine learning models (such as Random Forest \cite{breiman2001random}) for a user churn problem in the telecom industry \cite{idris2012churn}. 

In this paper, the mRMR method is studied and evaluated for implementation in an automated machine learning platform (Auto ML) for marketing. 
The goal of this platform is to democratize machine learning techniques for marketers and provide an easy and scalable solution for intelligent customer acquisition, re-engagement, cross-sell, up-sell, etc. The platform consists of independent training and prediction pipelines and the training pipeline consists of several modules such as data extraction, feature engineering, feature selection, model training etc. The best performing feature selection method, as discussed in the evaluation section of this paper, is implemented at scale in the Auto ML platform described in the implementation section of the paper.

The contribution of this paper is in the following areas:
\begin{itemize}
    \item Proposing two extensions to the mRMR framework with non-linear association for redundancy measure and model-based feature importance for relevance measure. 
    \item Performing extensive empirical evaluation of different variants of mRMR methods with one synthetic dataset and three real-world marketing datasets covering different business use cases. Both model performance and computation speed are evaluated, and the best mRMR variant is chosen for implementation at scale. 
    \item Describing the implementation and deployment of the mRMR method in a large-scale automated machine learning platform for marketing purpose. 
\end{itemize}

This paper is organized as follows: Section 
\ref{sec-background} introduces the existing mRMR framework and different variants. Section \ref{sec-extension} proposes two extensions to the mRMR framework, that leads to three new feature selection methods. Section \ref{sec-evaluation} evaluates different feature selection methods in both synthetic dataset and three real-world marketing datasets. Section \ref{sec-implementation} describes the implementation of the feature selection method in the automated machine learning platform. Finally, Section \ref{sec-conclusion} concludes the study and discusses potential extensions. 

\section{Background and Related Work}
\label{sec-background}

\subsection{mRMR Framework}
Assuming there are in total $m$ features, and for a given feature $X_i$ ($i \in \{1,2,...,m\}$), its feature importance based on the mRMR criterion can be expressed as \cite{peng2005feature}:
\begin{eqnarray}
\label{formula-mid}
f^{mRMR}(X_i) = I(Y, X_i) - \frac{1}{|S|} \sum_{X_s \in S} I(X_s, X_i)
\end{eqnarray}
where $Y$ is the response variable (class label), $S$ is the set of selected features, $|S|$ is the size of the feature set (number of features), $X_s \in S$ is one feature out of the feature set $S$, $X_i$ denotes a feature currently not selected: $X_i \not\in S$. The function $I(\cdot, \cdot)$ is the mutual information:
\begin{eqnarray}
I(Y,X) = \int_{\Omega_Y} \int_{\Omega_X} p(x,y) \log(\frac{p(x,y)}{p(x)p(y)}) dx dy
\end{eqnarray}
where $\Omega_Y$ and $\Omega_X$ are the sample spaces corresponding to $Y$ and $X$, $p(x,y)$ is the joint probability density, and $p(\dot)$ is the marginal density function. 

For discrete variables $Y$ and $X$, the mutual information formula takes the form:
\begin{eqnarray}
I(Y,X) = \sum_{y \in \Omega_Y} \sum_{x \in \Omega_X} p(x,y) \log(\frac{p(x,y)}{p(x)p(y)}).
\end{eqnarray}

In the mRMR feature selection process, at each step, the feature with the highest feature importance score $\max_{X_i \not\in S} f^{mRMR}(X_i)$ will be added to the selected feature set $S$. 

\subsection{Different mRMR Variants}
Under the mRMR framework, four variants are discussed in \cite{ding2005minimum}. 

For discrete features, the MID (mutual information difference) indicates the original form specified in formula (\ref{formula-mid}).
\begin{eqnarray}
f^{MID}(X_i) = I(Y, X_i) - \frac{1}{|S|} \sum_{X_s \in S} I(X_s, X_i).
\end{eqnarray}

As the MID uses the difference to balance the relevance and redundancy, the MIQ (mutual information quotient) uses a quotient scheme:
\begin{eqnarray}
f^{MIQ}(X_i) = I(Y, X_i) / [\frac{1}{|S|} \sum_{X_s \in S} I(X_s, X_i)].
\end{eqnarray}

For continuous features, as estimating the probability density can be computationally expensive, two alternatives are proposed in \cite{ding2005minimum}. 

The FCD (F-test correlation difference) uses the F-statistic to score the relevance, and correlation to score the redundancy:
\begin{eqnarray}
f^{FCD}(X_i) = F(Y, X_i) - \frac{1}{|S|} \sum_{X_s \in S} \rho(X_s, X_i),
\end{eqnarray}
where $\rho(X_s, X_i)$ is the Pearson correlation, and $F(Y, X_i)$ is the F-statistic. 

Similarly, the FCQ (F-test correlation quotient) uses the quotient scheme: 
\begin{eqnarray}
f^{FCQ}(X_i) = F(Y, X_i) / [\frac{1}{|S|} \sum_{X_s \in S} \rho(X_s, X_i)].
\end{eqnarray}

\section{Proposed mRMR Extensions}
\label{sec-extension}
In this section, two extensions to the mRMR method are made. 
As many features has non-linear association with the response variable in practice, one extension is to replace the Pearson correlation for redundancy in FCQ and FCD to a non-linear association measure called RDC (randomized dependence coefficient) \cite{lopez2013randomized}. 
On the other hand, if the downstream machine learning model is known, a relevance measure related to the model can be helpful. This leads to the other extension to replace the F-statistic with the embedded feature importance score, such as random forests feature importance score. 

\subsection{Non-linear Association Extension for Redundancy} 
The RDC developed by \cite{lopez2013randomized} measures the non-linear dependence between two random variables, as the largest canonical correlation between multiple randomly chosen nonlinear projections of their copula transformations. 
While the details can be found in the original paper \cite{lopez2013randomized}, the RDC algorithm can be summarized as the following steps:
\begin{itemize}
    \item Copula transformation of the variable (response variable and feature):
    \item Projecting the transformed features through various randomly chosen nonlinear kernels.
    \item Calculating the canonical correlation between the two sets of transformed variables, and using the largest correlation as the final RDC score. 
\end{itemize}

With RDC score, a new mRMR method FRQ 
(F-test RDC Quotient) criterion can be defined as:
\begin{eqnarray}
f^{FRQ}(X_i) = F(Y, X_i) / [\frac{1}{|S|} \sum_{X_s \in S} \rho_{rdc}(X_s, X_i)].
\end{eqnarray}

\subsection{Model-based Feature Importance Extension for Relevance} 
in the marketing applications considered in this paper's scope, the frequently used models are the tree-based classification models such as random forests (\cite{breiman2001random}) and gradient boosting tree (\cite{friedman2002stochastic, chen2016xgboost}). With such prior knowledge about the model, it is of interest to evaluate whether such information can contribute to the feature selection stage. An extension is made to mRMR method by using a model embedded feature importance score as the relevance measure in the mRMR method. 

One challenge of not using the same measure between relevance and redundancy is that the scales may be different between these two, that affects the trade-off between the relevance and redundancy. Compared with the difference scheme (such as MID, FCD), the quotient scheme (such as MIQ, FCQ) is relatively immune to the scale differences. As the proposed model-based feature importance measure for relevance has different scale from the redundancy measure (i.e. correlation), the quotient scheme is adopted instead of the difference scheme.

The random forest feature importance score (\cite{menze2009comparison}) is denoted as $I_{RF}(Y, X_i)$. The proposed extension RFCQ (Random-Forest correlation quotient) is defined as:
\begin{eqnarray}
f^{RFCQ}(X_i) = I_{RF}(Y, X_i) / [\frac{1}{|S|} \sum_{X_s \in S} \rho(X_s, X_i)].
\end{eqnarray}

Similarly, the RFRQ (Random-Forest RDC quotient) can be defined as:
\begin{eqnarray}
f^{RFRQ}(X_i) = I_{RF}(Y, X_i) / [\frac{1}{|S|} \sum_{X_s \in S} \rho_{rdc}(X_s, X_i)].
\end{eqnarray}

\section{Empirical Evaluation}
\label{sec-evaluation}
In order to select the best-performing mRMR method for application and platform implementation, an empirical study is performed to evaluate the performance of the mRMR methods on both synthetic data and real data sets. Three real marketing data sets are used in this evaluation, covering multiple business use cases using classification models. 

\subsection{Setup}
The empirical study on different data sets share the same evaluation setup, algorithm implementation, and computation environment.

The mRMR variants discussed above are implemented in Python and evaluated in this study, including MID, MIQ, FCD, FCQ, FRQ, RFCQ, and RFRQ. 
In addition, the Random Forest feature importance (RF as abbreviation) is used as a benchmark feature selection method that is not within the mRMR family. For the Random Forest feature importance method, the feature is ranked by the feature importance generated by the Random Forest classifier and the top features can be selected according to the feature importance rank. Another benchmark is using all the features in the classification model without any feature selection. 
The feature selection methods used in the evaluation are summarized in Table \ref{table:mrmr-methods}.

\begin {table}
\begin{center}
\caption{Feature Selection Methods in the Empirical Study}
\label{table:mrmr-methods}
\begin{tabular}{|c c c c|}
    \hline
    \textbf{Method} & \textbf{Relevance} & \textbf{Redundancy} & \textbf{Scheme}  \\ [0.5ex] 
    \hline
    MID             & Mutual        & Mutual      & Difference \\%
                    & Information   & Information  &(Relevance - Redundancy) \\
    \hline
    MIQ             & Mutual        & Mutual      & Quotient \\%
                    & Information   & Information & (Relevance / Redundancy) \\
    \hline
    FCD             & F Statistic        & Correlation         & Difference \\
    \hline
    FCQ             & F Statistic        & Correlation         & Quotient \\
    \hline
    FRQ             & F Statistic        & RDC                 & Quotient \\
    \hline
    RFCQ            & Random Forests     & Correlation         & Quotient \\
    \hline
    RFRQ            & Random Forests     & RDC                 & Quotient \\
    \hline
    RF            & Random Forests     & N/A                 & N/A \\
    \hline
\end{tabular}
\end{center}
\end{table}

The goal of the empirical evaluation is to select the feature selection method with good and robust practical performance across various scenarios such that it can be implemented at scale as a platform solution. 
The performance is evaluated in two dimensions. First, the computation speed is measured for each feature selection method for ranking top features. Second, the accuracy metrics are measured through the downstream classification models using the features set selected by each feature selection method. The model accuracy metrics include AUC (Area Under the Curve) and F1-score \cite{powers2011evaluation}. Three downstream classification models are used to evaluate the performance using the selected features: Naive Bayes \cite{rish2001empirical} as probabilistic classifier, Logistic Regression \cite{hosmer2013applied} as regression-based classifier, and Random Forest classifier \cite{breiman2001random} as tree-based classifier. As these three classification models come from different classification model families, the evaluation on feature selection method can be representative for the general classification models, instead of a specific model. The hyper-parameters for the Random Forest classifier (applicable for both the feature selection method and classification model where the Random Forest is used) are set as: $50$ for number of trees, $10$ for tree depth, $50$ for minimum samples in the leaf node, and entropy as the split loss function.

For each data set, the evaluation procedure is carried out according to the following steps:
\begin{itemize}
  \item Step 1: Dividing the data set into training data and testing data by random splitting. For synthetic data, a new data set is generated in each iteration of $10$ total data generation trials, and the split only happens once for each data set ($50\%$ for training and $50\%$ for testing). For real data, four-fold cross-validation \cite{kohavi1995study} is used ($75\%$ for training and $25\%$ for testing at each cross-validation iteration). 
  \item Step 2: Running each feature selection method on the training data to select and rank the top features (top $20$ for synthetic data and top $30$ for real data) out of all features. 
  \item Step 3: Fitting each classification model on the training data using the top $m$ features for each $m=1,2,...,20$ from the ranked feature set of each feature selection method. 
  \item Step 4: Predicting the label probabilities on the testing data using the fitted classification models. 
  \item Step 5: Measuring the classification model accuracy based on the predicted results and the true labels in the testing data. Such performance metrics (AUC, F1-Score) are averaged over the iterations for data generation trials of synthetic data and cross-validation of real data. 
\end{itemize}

In sum, this evaluation setup leads to $8$ (feature selection methods) $\times$ $3$ (classification models) $\times$ $4$ (data sets) $ = 96$ sets of empirical performance results.

\subsection{Synthetic Data Example}

\begin{figure}
\centering
\includegraphics[width=0.4\textwidth]{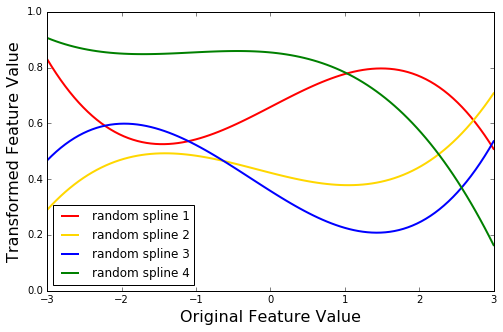}
\caption{Example Splines Used for Generating Nonlinear Feature Associations for Synthetic Data}
\label{fig:simulation_splines}
\end{figure}

\begin{figure*}[h]
\includegraphics[width=1\textwidth]{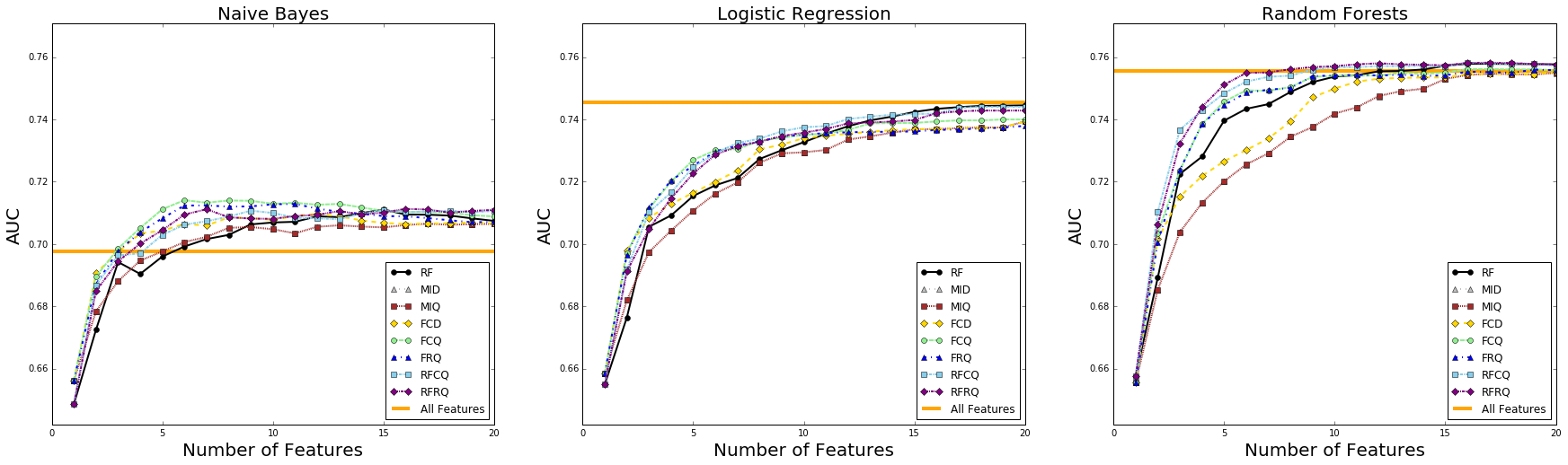}
\includegraphics[width=1\textwidth]{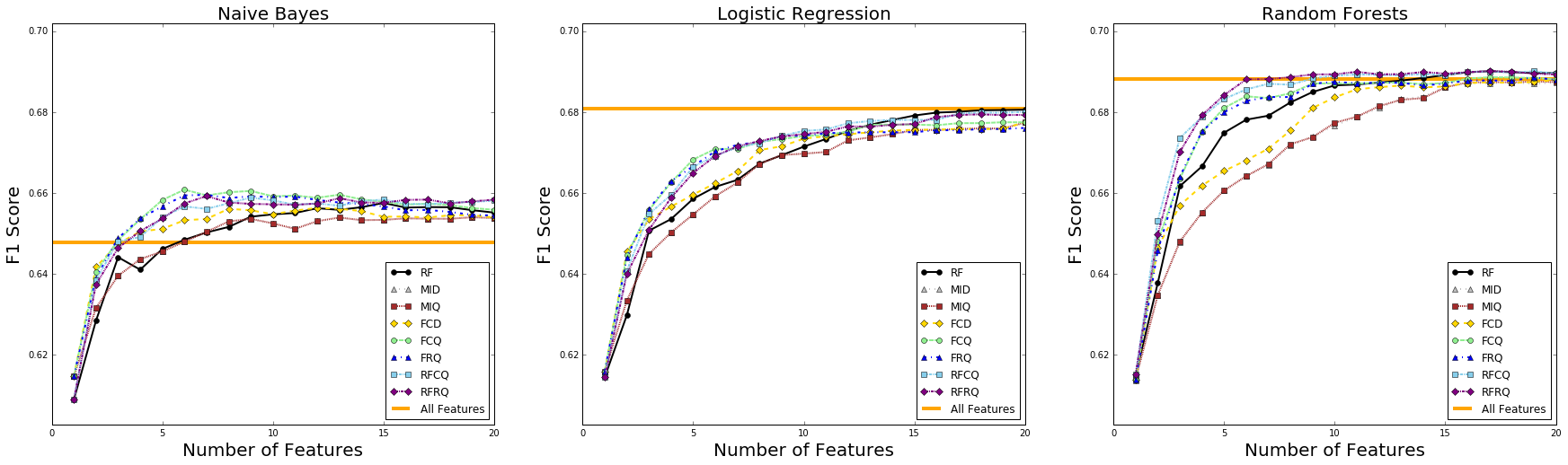}
\caption{Synthetic Data AUC (top row) and F1 Score (bottom row) for Different Feature Selection Methods and Classification Models}
\label{fig:synthetic_results}
\end{figure*}

\begin{figure*}[h]
\centering
\includegraphics[width=1\textwidth]{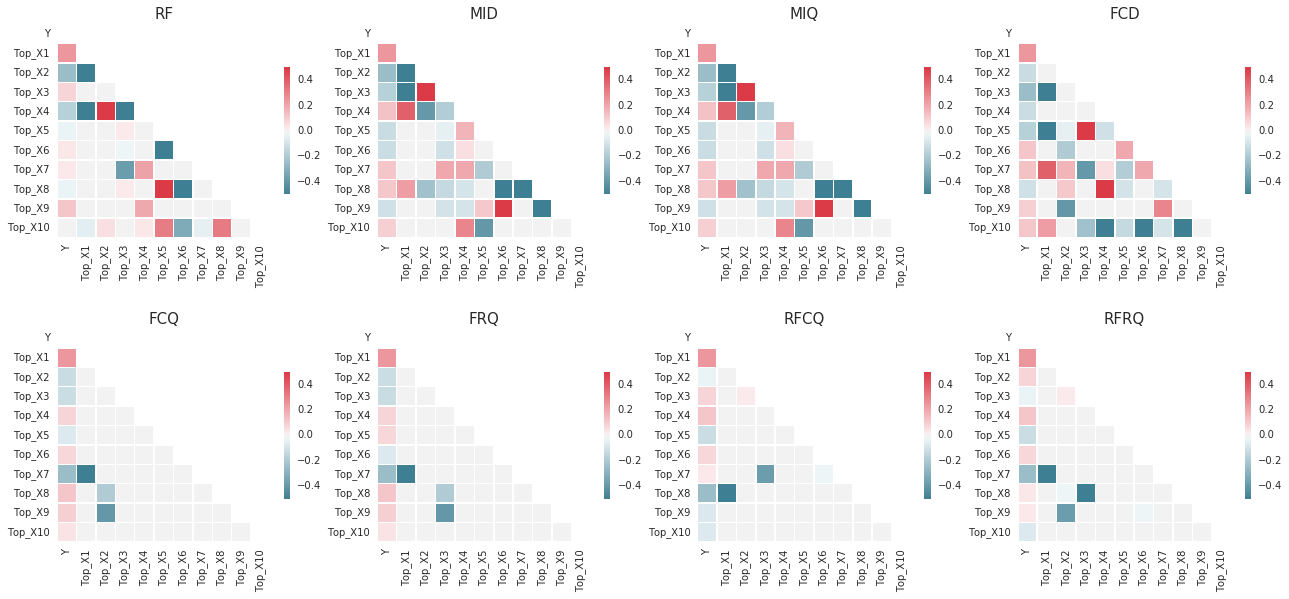}
\caption{Synthetic Data Correlation Heatmap among Top Features Selected by Each Methods and Response Variable $Y$ : the good-performing mRMR methods (FCQ, FRQ, RFCQ, RFRQ) effectively reduce the redundancy among the top features.}
\label{fig:synthetic_corr}
\end{figure*}

\begin {table*}
\begin{center}
\caption{Synthetic Data Evaluation in AUC Scores \scriptsize (The highest scores within each column are highlighted.)}
\label{table:synthetic_auc_f1}
\begin{tabular}{|c| c c c| c c c| c c c|}
\hline
Model &  \multicolumn{3}{c|}{Naive Bayes}   & \multicolumn{3}{c|}{Logistic Regression}   & \multicolumn{3}{c|}{Random Forests}\\
\hline
$\#$ Features & 5 & 10 & 15 & 5 & 10 & 15 & 5 & 10 & 15 \\
\hline
MID & 0.698 & 0.705 & 0.705 & 0.711 & 0.729 & 0.737 & 0.720 & 0.742 & 0.753 \\
MIQ & 0.698 & 0.705 & 0.705 & 0.711 & 0.729 & 0.737 & 0.720 & 0.742 & 0.753 \\
FCD & 0.704 & 0.708 & 0.707 & 0.716 & 0.734 & 0.737 & 0.727 & 0.750 & 0.753 \\
FCQ & \textbf{0.711} & \textbf{0.713} & \textbf{0.711} & \textbf{0.727} & 0.735 & 0.739 & 0.746 & 0.754 & 0.755 \\
FRQ & 0.708 & \textbf{0.713} & 0.709 & 0.725 & 0.735 & 0.736 & 0.745 & 0.754 & 0.754 \\
RFCQ & 0.703 & 0.710 & \textbf{0.711} & 0.725 & \textbf{0.737} & \textbf{0.742} & 0.748 & \textbf{0.757} & \textbf{0.757} \\
RFRQ & 0.704 & 0.708 & 0.710 & 0.723 & 0.736 & 0.740 & \textbf{0.751} & \textbf{0.757} & \textbf{0.757} \\
RF & 0.696 & 0.707 & \textbf{0.711} & 0.715 & 0.733 & \textbf{0.742} & 0.740 & 0.754 & \textbf{0.757} \\
\hline
All Features (70) & \multicolumn{3}{c|}{0.698}   & \multicolumn{3}{c|}{0.746}   & \multicolumn{3}{c|}{0.756}  \\
\hline
\end{tabular}
\end{center}
\end{table*}

\begin {table}
\begin{center}
\caption{Computation Time (in Seconds) for Selecting and Ranking Top Features}
\label{table:synthetic_time}
\begin{tabular}{|c|c|c|c|c|}
\hline 
Method & Synthetic Data & Dataset 1 & Dataset 2 & Dataset 3 \\
\hline 
MID & 187 & 2593 & 831 & 2732 \\
MIQ & 187 & 2380 & 743 & 1780 \\
FCD & 1 & 31 & 11 & 17 \\
FCQ & 1 & 31 & 11 & 16 \\
FRQ & 276 & 4699 & 1993 & 1204 \\
RFCQ & 60 & 45 & 27 & 241 \\
RFRQ & 331 & 1567 & 1209 & 1289 \\
RF & 60 & 41 & 23 & 429 \\
\hline
\end{tabular}
\end{center}
\end{table}

The synthetic data is generated for a classification problem. Compared with the real data, the advantage of the synthetic data is that it is controlled and known that how many features are informative and redundant, as well as the real relationship between feature and label, and among features. 

There are four types of features in the synthetic data generation: (1) independent informative features as features influencing the conversion probability; (2) the linear redundant features as a linear combination of a random subset of independent informative features; (3) the nonlinear redundant features as a nonlinear combination of a random subset of independent informative features; (4) the irrelevant features that are independent of conversion probability and any other features. 

The synthetic data is generated in following steps:
\begin{itemize}
\item Generating the labeled data with a logistic regression model:
\begin{eqnarray}
p_i = \frac{1}{1+exp(-\sum_{j=1}^{10} (z_{ij} \beta_j + e_{ij})}
\end{eqnarray}
where $p_i$ is the conversion probability for the $i$th user ($i=1,2,...,n$ with $n=10^5$ in this case), $z_{ij}$ is the $j$th ($j=1,2,...,10$) covariate for the $i$th user generated independently from standard normal distribution $N(0,1)$, $\beta_j$ is the covariate coefficient generated from a uniform distribution $U(-1,1)$, and $e_{ij}$ is an error term generated independently from a normal distribution with $0$ mean and $0.1$ standard deviation: $N(0,0.1)$. The binary label $Y_i$ is then generated based on $p_i$: $Y_i = 1$ if $p_i >= 0.5$, otherwise $Y_i = 0$.

\item Generating independent informative features by a nonlinear transformation of $z_{ij}$. The nonlinear transformation functions are splines \cite{reinsch1967smoothing} created by fitting a spline to $10$ random points $(w_k, v_k)$ for $w_k$ as equal spaced points between $[-3,3]$ and $v_k$ as uniform random variables: $v_k \sim U(0,1)$. Example splines can be found in Figure \ref{fig:simulation_splines}. There are in total $10$ random spline functions generated using this procedure. The independent informative feature $x_{ij}$ are generated by applying a randomly selected spline function $f(\cdot)$ on $z_{ij}$: $x_{ij} = f(z_{ij})$
Note that after the nonlinear transformation, the independent informative feature $x_{ij}$ also has a nonlinear relationship with the conversion probability odds  (closer to real scenario).

\item Creating the linear redundant features as a linear combination of a random subset of independent informative features. The random feature subset is created by first randomly sampling a number from $1$ to $10$ with equal probability, and then sampling the corresponding number of independent informative features. The coefficients in the linear combination is generated independently from the standard normal distribution.

\item Creating the nonlinear redundant feature by first generating a intermediate linear redundant feature and then apply a randomly selected spline function $f(\cdot)$ on the intermediate linear redundant feature. 

\item Finally, generating the irrelevant features by randomly sampling from the standard normal distribution $N(0,1)$.
\end{itemize}

In this synthetic data set, there are $70$ features in total, including $10$ independent informative features, $20$ linear redundant features, $20$ nonlinear redundant features, and $20$ irrelevant features.

The evaluation results based on the synthetic data are shown in Figure \ref{fig:synthetic_results} and Table \ref{table:synthetic_auc_f1}. 

In Figure \ref{fig:synthetic_results}, both AUC and F1 Score are shown, and the relative performance of different methods is similar between AUC and F1 Score. Therefore, the remaining discussion is mainly around AUC, but the conclusion can be generalized to F1 Score. 

\begin{figure*}[h]
\includegraphics[width=1\textwidth]{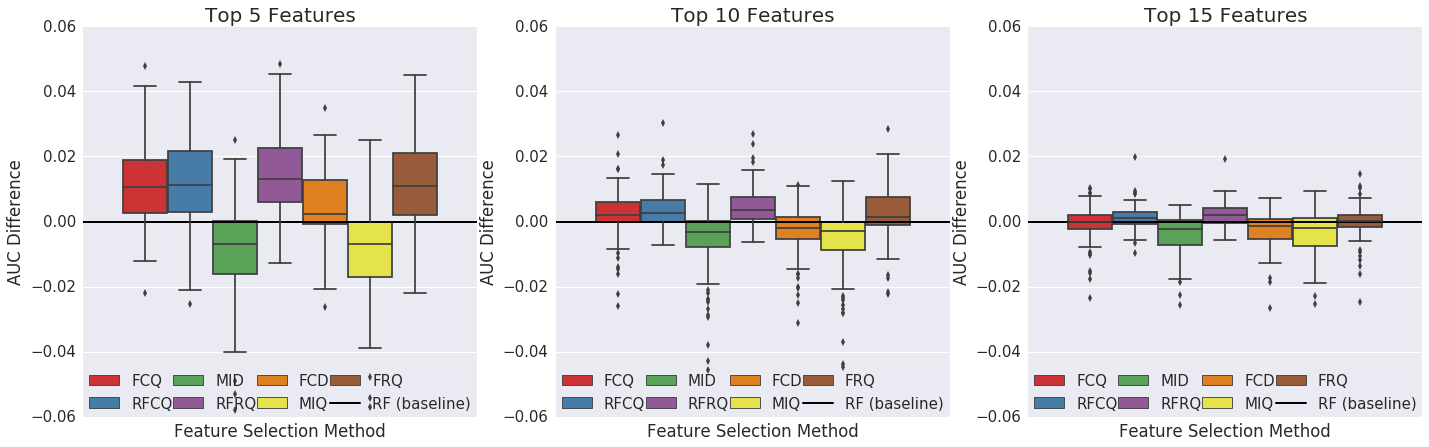}
\caption{AUC Gain compared with RF Feature Selection method as a baseline, in Random Forest Classifier. The results are based on $100$ trials of synthetic data simulation. In general, the mRMR methods showed advantage over the RF method especially in small feature size scenarios. The RFCQ and RFRQ methods showed advantages in all three scenarios.}
\label{fig:auc_difference}
\end{figure*}

Comparing across different classification models, in general, the Random Forest model outperforms the other two models with the same set of features. For the Naive Bayes model, there is an obvious over-fitting issue by using all the $70$ features, and the feature selection methods improve the performance by building the model on a selected feature subset. For the other two classification models, the benefits of using the feature selection methods is to achieve a similar model performance as using all the features with a much compact feature set. For example, the RFRQ method can boost the Random Forest performance to the optimal with as few as $6$ features. 
As there are $10$ independent informative features, the expectation is that the top $10$ features selected by a good feature selection method can yield good model performance, and including additional features beyond $10$ will produces minimal incremental performance gain. This pattern can be observed for Naive Bayes and Random Forest. However, the performance of Logistic Regression keeps improving after $10$ features, and a potential reason is that the additional nonlinear redundant features can provide additional nonlinear information to the regression model. 

The relative performance by different feature selection methods can be seen in the Table \ref{table:synthetic_auc_f1}. The AUC scores are shown for different classification models and feature selection methods for top $5$, $10$, $15$ features. The best performing feature selection method under each classification model and feature size combination is highlighted. In general, FCQ, RFCQ, and RFRQ are performing well. The RF method, without considering redundancy, begins to show good performance with $15$ features, while RFCQ and RFRQ can achieve such performance with less features. As a comparison, the performance of MID, MIQ, and FCD are relatively poor. The reason for MID and MIQ methods are the difficulty and accuracy for estimating the empirical probability distribution of the features. And the potential explanation for FCD method is the scale imbalance between the F statistic as the relevance measure and the correlation as the redundancy measure in the difference scheme. 

In this synthetic data, the extension of mRMR framework to model-based feature importance score (RFCQ, RFRQ) seems to be successful, while the extension from linear correlation to nonlinear association using RDC method does not show significant improvement. 

Figure \ref{fig:synthetic_corr} show the correlation between top features selected by different models as well as the correlation between the features and the conversion ($Y$ variable). The top features selected by RF are highly correlated, while the top features selected by the mRMR methods (FCQ, RFCQ, RFRQ) are less correlated, especially for the top $6$ features. This figure visualizes the effect of reducing redundancy by mRMR methods. Such a redundancy reduction leads to a better classification model performance discussed above, especially when the selected feature subset is small in size. In addition, such distinctive features would also provide better model insights and interpretation in practice, without overwhelming the reader with redundant information. 

The computation speed metrics are recorded in Table \ref{table:synthetic_time}. The FCD and FCQ methods are the fastest methods, followed by RF and RFCQ methods, and the rest methods MID, MIQ, FRQ, RFRQ are relatively computationally expensive. 

To further examine the statistical significance on performance of different feature selection methods, a more extensive study is performed with $100$ iterations of simulations. Except reducing the sample size to $10,000$ for each trial to accelerate the computation time, all other settings are kept the same as before. The results are shown in Figure \ref{fig:auc_difference}. The AUC is calculated for each trial for each feature selection method using the Random Forest classifier, and the AUC difference is calculated as the difference between the AUC from a given feature selection method and the AUC from RF feature selection method as a baseline. Such AUC difference is calculated for each trial, and the distribution of the AUC differences is plotted as a box-plot in the figure. As the classifier is Random Forest, there is natural advantage of the RF feature selection method. Under this scenario, the FCQ, RFCD, and RFRQ methods perform better than the RF method, especially when there are a small set of features selected (Top $5$ Features or Top $10$ Features). 

Overall, in the synthetic data evaluation, the FCQ and RFCQ methods show good performance in both accuracy metrics and computation time metrics.

\subsection{Real Data Examples}

\begin {table*}
\begin{center}
\caption{Meta Information about Real Datasets}
\label{table:real_datasets}
\begin{tabular}{|c|c|c|c|}
\hline 
Real Dataset & Use Case & Number of Features & Number of Users \\
\hline 
Dataset 1 & Cross-sell product A to existing users of product B within the same app & 1300 & $10^5$ \\
Dataset 2 & Up-sell product A to existing users of product B within the same app & 570 & $10^5$ \\
Dataset 3 & Cross-sell one mobile app A to existing users of another app B & 100 & $10^6$ \\
\hline

\end{tabular}
\end{center}
\end{table*}

\begin{figure*}
\centering
\textbf{Dataset 1 \\}
\includegraphics[width=1\textwidth]{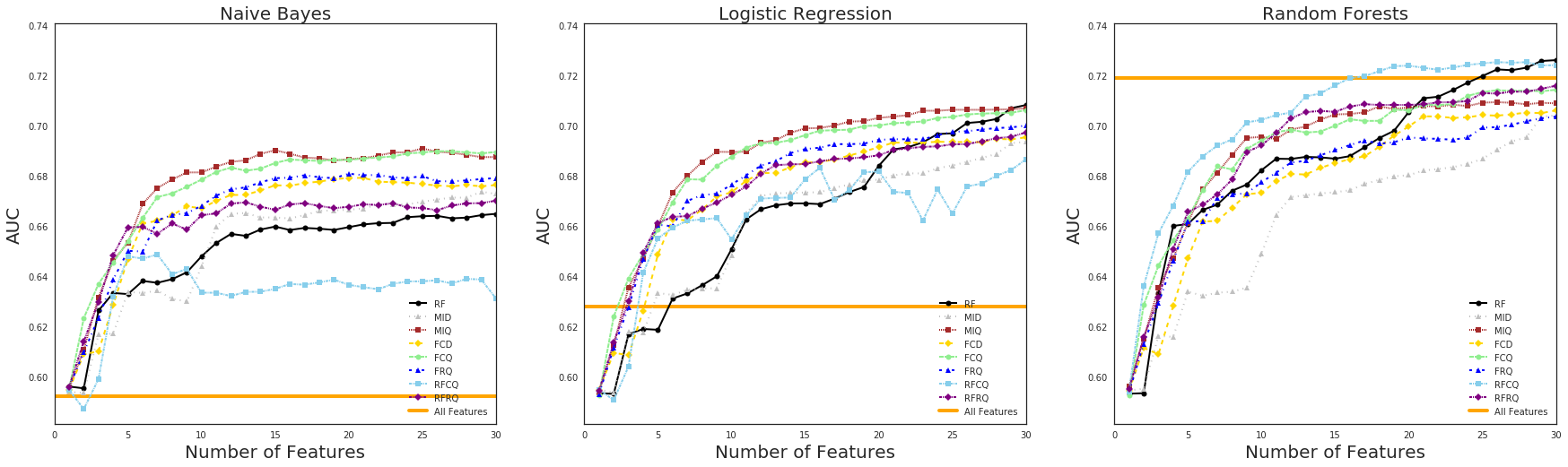}
\textbf{Dataset 2 \\}
\includegraphics[width=1\textwidth]{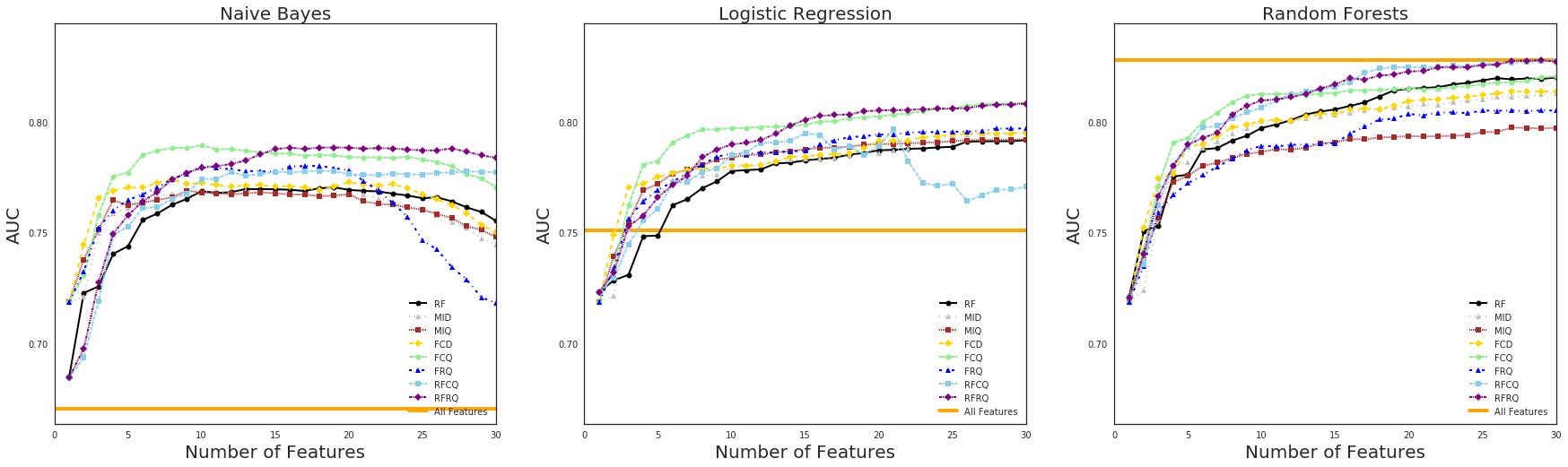}
\textbf{Dataset 3 \\}
\includegraphics[width=1\textwidth]{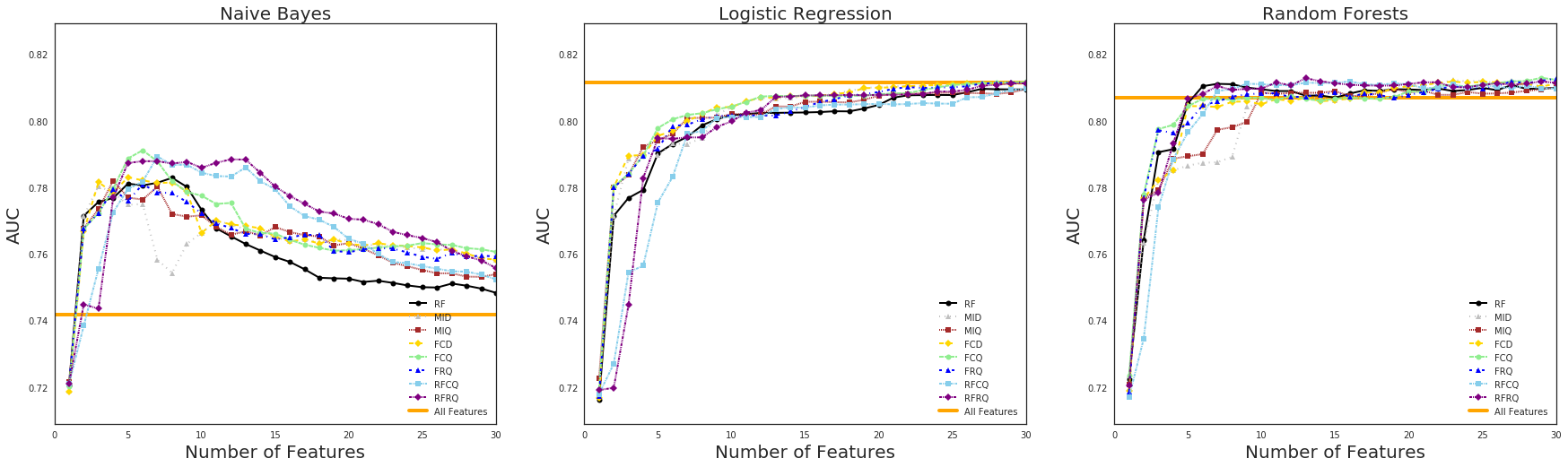}
\caption{Real Data AUC Score for Different Feature Selection Methods and Classification Models (The rows of the plots are sorted in order of Dataset 1, 2, and 3.}
\label{fig:real_data_auc}
\end{figure*}

\begin {table*}
\begin{center}
\caption{Real Data Evaluation in AUC Scores \scriptsize (The highest scores within each column are highlighted.)}
\label{table:real_data_auc}
\begin{tabular}{|c| c c c| c c c| c c c|}
\hline
Model &  \multicolumn{3}{c|}{Naive Bayes}   & \multicolumn{3}{c|}{Logistic Regression}   & \multicolumn{3}{c|}{Random Forests}\\
\hline
$\#$ Features & 10 & 20 & 30 & 10 & 20 & 30 & 10 & 20 & 30 \\
\hline
\multicolumn{10}{|c|}{Dataset 1} \\
\hline
MID & 0.644 & 0.667 & 0.673 & 0.649 & 0.679 & 0.694 & 0.649 & 0.681 & 0.704 \\ 
MIQ & \textbf{0.682} & \textbf{0.687} & 0.688 & \textbf{0.690} & \textbf{0.704} & 0.707 & 0.696 & 0.708 & 0.709 \\  
FCD & 0.667 & 0.680 & 0.677 & 0.674 & 0.692 & 0.696 & 0.674 & 0.700 & 0.706 \\ 
FCQ & 0.679 & \textbf{0.687} & \textbf{0.690} & 0.688 & 0.700 & 0.707 & 0.694 & 0.707 & 0.715 \\ 
FRQ & 0.668 & 0.681 & 0.680 & 0.677 & 0.695 & 0.700 & 0.678 & 0.696 & 0.704 \\ 
RFCQ & 0.634 & 0.637 & 0.632 & 0.655 & 0.682 & 0.687 & \textbf{0.703} & \textbf{0.724} & 0.725 \\ 
RFRQ & 0.665 & 0.668 & 0.671 & 0.673 & 0.689 & 0.698 & 0.693 & 0.709 & 0.716 \\ 
RF & 0.648 & 0.660 & 0.665 & 0.651 & 0.684 & \textbf{0.709} & 0.683 & 0.706 & \textbf{0.727} \\ 
\hline
All Features (1300) & \multicolumn{3}{c|}{0.593}   & \multicolumn{3}{c|}{0.628}   & \multicolumn{3}{c|}{0.719}  \\
\hline
\multicolumn{10}{|c|}{Dataset 2} \\
\hline
MID & 0.769 & 0.768 & 0.745 & 0.780 & 0.786 & 0.795 & 0.801 & 0.807 & 0.814 \\ 
MIQ & 0.768 & 0.768 & 0.748 & 0.784 & 0.790 & 0.793 & 0.787 & 0.794 & 0.798 \\ 
FCD & 0.773 & 0.773 & 0.751 & 0.781 & 0.792 & 0.796 & 0.801 & 0.810 & 0.814 \\ 
FCQ & \textbf{0.790} & 0.785 & 0.771 & \textbf{0.798} & 0.803 & \textbf{0.809} & \textbf{0.813} & 0.816 & 0.821 \\ 
FRQ & 0.781 & 0.779 & 0.719 & 0.786 & 0.795 & 0.797 & 0.790 & 0.804 & 0.806 \\ 
RFCQ & 0.775 & 0.777 & 0.778 & 0.785 & 0.790 & 0.771 & 0.807 & \textbf{0.825} & \textbf{0.828} \\ 
RFRQ & 0.780 & \textbf{0.789} & \textbf{0.784} & 0.790 & \textbf{0.806} & \textbf{0.809} & 0.810 & 0.823 & \textbf{0.828} \\ 
RF & 0.769 & 0.770 & 0.756 & 0.778 & 0.788 & 0.792 & 0.798 & 0.815 & 0.821 \\ 
\hline
All Features (570) & \multicolumn{3}{c|}{0.671}   & \multicolumn{3}{c|}{0.752}   & \multicolumn{3}{c|}{0.828}  \\
\hline
\multicolumn{10}{|c|}{Dataset 3} \\
\hline
MID & 0.767 & 0.764 & 0.759 & 0.804 & \textbf{0.811} & \textbf{0.812} & 0.806 & \textbf{0.811} & 0.812 \\ 
MIQ & 0.772 & 0.764 & 0.754 & 0.802 & 0.808 & 0.810 & 0.809 & 0.809 & 0.810 \\ 
FCD & 0.767 & 0.763 & 0.759 & 0.804 & 0.810 & \textbf{0.812} & 0.805 & \textbf{0.811} & 0.811 \\ 
FCQ & 0.778 & 0.761 & \textbf{0.761} & \textbf{0.805} & 0.808 & \textbf{0.812} & 0.807 & 0.809 & \textbf{0.813} \\ 
FRQ & 0.772 & 0.761 & 0.760 & 0.802 & 0.809 & \textbf{0.812} & 0.809 & 0.808 & 0.813 \\ 
RFCQ & 0.785 & 0.765 & 0.753 & 0.801 & 0.805 & 0.810 & \textbf{0.811} & \textbf{0.811} & 0.810 \\ 
RFRQ & \textbf{0.786} & \textbf{0.771} & 0.756 & 0.800 & 0.808 & \textbf{0.812} & 0.810 & \textbf{0.811} & 0.812 \\ 
RF & 0.774 & 0.753 & 0.749 & 0.802 & 0.805 & 0.810 & 0.810 & 0.810 & 0.810 \\ 
\hline
All Features (100)  & \multicolumn{3}{c|}{0.742}   & \multicolumn{3}{c|}{0.812}   & \multicolumn{3}{c|}{0.807}  \\
\hline
\end{tabular}
\end{center}
\end{table*}

\begin {table}
\begin{center}
\caption{Hyperparameter Variation for Random Forest Classifier on Dataset 2 \scriptsize (The highest scores within each column are highlighted. Different hyperparameter setting is used for each Random Forest model.)}
\label{table:hyperp}
\begin{tabular}{|c|c|c|}
\hline
Model &  \multicolumn{1}{c|}{Random Forest v1}   & \multicolumn{1}{c|}{Random Forest v2} \\
\hline
$\#$ Features Selected & 5 & 5 \\
\hline
MID & 0.784	& 0.779 \\ 
MIQ &  0.777 & 0.776 \\  
FCD &  0.789 &	0.784\\ 
FCQ &  0.791 & 0.787\\ 
FRQ & 0.767 & 0.766 \\ 
RFCQ & 0.79 & 0.786 \\ 
RFRQ & \textbf{0.793} & \textbf{0.789} \\ 
RF & 0.777 & 0.772\\ 
\hline
\end{tabular}
\end{center}
\end{table}

Three real-world marketing-related datasets at Uber are used to empirically evaluate the feature selection methods for predicting the likelihood of a user taking a particular action (e.g. using another product). Uber has multiple product offerings, including ride-sharing service (UberX, Uber Black, Uber Pool, etc.), food delivery service (Uber Eats), as well as Bike and Scooter rentals. In addition, there are also region-specific products launched in different countries. In order to encourage existing users of one product to explore other products, marketing efforts are made to identify potential users who are likely to try these products. These datasets, as described in Table \ref{table:real_datasets}, range across a number of different use cases and span across multiple regions in the world (such as United States, Latin America, and India). These datasets are standardized at a user level, with $100$ thousand to $1$ million observations and $100 \sim 1000$ features consolidated from multiple channels within the business. 
The user level features range from static features (such as geographic location) to dynamic app usage frequency features (such as number of orders in the past one week). In addition to user-level features, features defined at other levels include zip-code or city level and product type level to capture additional information. Dummy variables are created for categorical data.

The outcome AUC scores are summarized in Figure \ref{fig:real_data_auc} and Table \ref{table:real_data_auc}. The F1 Score patterns are consistent with the AUC patterns, as it can be seen in the synthetic example, therefore the F1 Score plots are not included for the real data example. 
Similar to the results in the synthetic data example, it can be observed that the AUC goes up rapidly with increasing number of features when the feature set is small ($<=5$). For Naive Bayes (in all cases) and Logistic Regression (for dataset 1 and 2), at a particular point the AUC scores with feature selection crosses above the orange line (as the AUC benchmark when all the features are included in the classification model). This suggests a potential case of overfitting when all the features are used in the model. 
If the x-axis is extended to look at the AUC score beyond just $30$ features, it is observed that the AUC score for the mRMR variants starts falling at some point to finally merge with the orange (all features) line. 

Within the three classification models, the Random Forests generally perform better than the other two models. Within Random Forests, the model-based feature selection methods, such as RFCQ and RFRQ, are performing well for selecting the important feature set. The model-free feature selection method FCQ also has a good performance that is close to optimal. For Logistic Regression and Naive Bayes, the model-based feature selection methods do not show a consistent advantage over other feature selection methods. On the other hand, FCQ as a model-free feature selection method shows a good performance across different datasets. 

Furthermore, Table \ref{table:hyperp} shows the AUC scores for hyperparameter variation within the Random Forest Classifier on Dataset 2. Using the notation from the sklearn package \cite{scikit-learn}, the hyperparameters for Random Forest v1 are: max\_features="sqrt" , max\_depth=10, min\_samples\_leaf=50, and n\_estimators=50. The hyperparameters for Random Forest v2 are: max\_features="log2", max\_depth=5 min\_samples\_leaf=20, and n\_estimators=20. By comparing these two hyperparameter settings, the Random Forest v1 performs better than the Random Forest v2. To compare different feature selection methods, RFRQ, RFCQ and FCQ consistently perform well for selecting the optimal feature set. 

For computation time, as shown in Table \ref{table:synthetic_time}, the FCD, FCQ, RFCQ, and RF methods run fairly quickly whereas the MID, MIQ, FRQ and RFRQ variants take much longer to run. The reasons contributing to relatively long running time are large feature set for the dataset 1 and large sample size for the dataset 3. In all scenarios evaluated, the FCQ and FCD methods take less than one minute to run. 

Overall, combining the accuracy measure and the computation efficiency measure, it can be seen that RFCQ can be a good choice if the down-stream classification model is Random Forests, but FCQ has an outstanding performance in computation time and robust accuracy for different downstream classification models. 

\section{Implementation in Production}
\label{sec-implementation}
The automated machine learning platform for marketing at Uber enables marketing teams to develop and deploy machine learning models in a quick, easy and scalable fashion. It also connects to the rich source of user data that Uber possesses to use for modeling. The platform is developed using Scala Spark and consists of separate pipelines for training and prediction.

In this platform, after running through the Feature Engineering and Transformation modules, a large set of features is generated that can be used for modeling. This is where the Feature Selection module comes in handy, to narrow down the most relevant and diversified subset of features to use for modeling. 
Based on the empirical evaluation results in Section \ref{sec-evaluation}, it was decided to implement the feature selection module in two phases. In phase $1$, the FCQ has been implemented in production considering its simplicity, robust performance, and fast computation speed. In phase $2$, more mRMR variants (such as RFCQ) will be added in production and the best performing algorithm will get selected dynamically, at run-time, according to model performance measure.

The following subsections dive deeper into the overall platform architecture, as well as the feature selector architecture.

\subsection{Architecture}
\begin{figure}
\vspace{0.2cm}
\centering
\includegraphics[width=8.5cm, height=6cm]{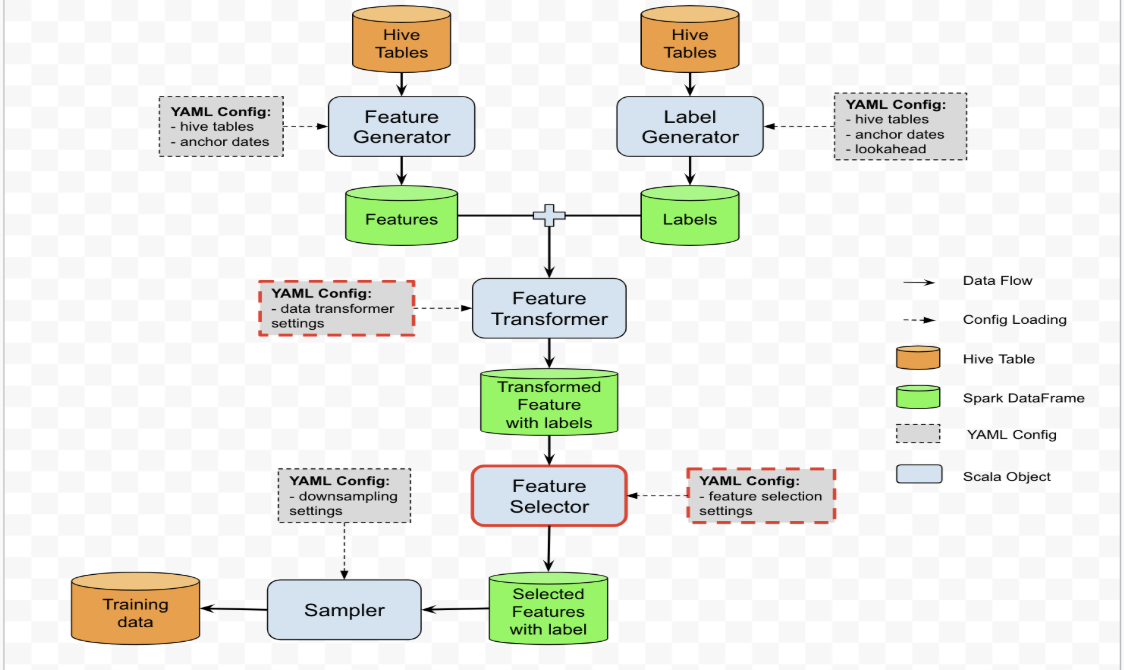}
\caption{Automated Machine Learning Platform Architecture (training pipeline)}
\label{fig:automl}
\end{figure}

Figure \ref{fig:automl} illustrates the architecture of the training pipeline which consists of several modules such as Feature and Label Generator, Feature Transformer, Feature Selector, Data Sampler and Model Training, in that order.

The Feature Selector module takes several inputs such as the dataframe with all the features and label, the feature selection method, the maximum number of features to select, as well as the pipeline type i.e. training or prediction. It outputs the dataframe with selected features, the list of selected features and the feature importance scores i.e. the mRMR scores.

The number of features to select is decided by running a Random Forest Model on the dataset. For every $n$ ranging from 1 to a sufficiently large number of features such as $n$ = 50, the AUC is calculated and the lowest $n$ with the highest AUC is chosen as the number of features to select. 

\subsection{Implementation}
The mRMR feature selection variants were first developed in Python and utilized for empirical evaluation, as described in the section above. To implement the module in production for the platform, Scala Spark was chosen as the medium of implementation. The decision was based on several factors such as the use of JVM at run-time which makes Scala faster than Python for large datasets, the use of multiple concurrency primitives allowing better memory management and data processing using Scala, as well as more powerful frameworks, libraries, implicit, macros etc. provided by Scala. Scala also interacts well with the upstream services which are implemented in Java.

\subsection{Challenges and Optimization}
The first version of the mRMR module took a considerably long time to run on the real datasets, which implied a need for further optimization and heuristic approximations to make it faster, while still retaining accuracy. This was done by:
\begin{itemize}
    \item Replacing loops with Scala’s map functionality.
    \item Making the best possible use of the Spark RDD and Dataframe APIs and using a combination of them to develop the most optimal code. While RDDs offer low-level functionality and control, dataframes allow custom view, high-level operations, save space, and run faster.
    \item Placing the feature selector module after the down-sampler module so instead of running on the entire dataset it runs on a smaller sample which is still representative of the dataset.
    \item Adding sophisticated error handling techniques and a default to keep all features if mRMR fails to run in time.
\end{itemize}

\subsection{Online Experiment Evaluation}
The cross-sell model for Dataset $3$ in Table \ref{table:real_datasets} was built and deployed through the automated machine learning platform with the FCQ feature selection method implemented. A campaign using this model for user targeting was tested online in a randomized experiment. Using this model, top $60\%$ (w.r.t. the predicted conversion probability) of treatment users were cross-sold a new product through app notifications and display channels. The experiment results proved that the model is effective in identifying the high conversion users: for example, the users with top $20\%$ predicted conversion probability in control have $4$ times higher actual new product adoption rate compared with baseline (all users in control). In addition, the marketing campaign creates $12\%$ incremental adoption for the top $60\%$ propensity users in treatment group compared with the equivalent top $60\%$ propensity users in the control group with statistical significance ($p < 0.05$). 

\section{Conclusion}
\label{sec-conclusion}
In this paper, different mRMR feature selection methods, including existing variants and proposed extensions, are empirically evaluated with both synthetic dataset and three real-world marketing datasets. In this study, the model-free FCQ variant shows robust performance for different classification models as well as high computation efficiency. On the other hand, the model-based variants RFCQ and RFRQ provide optimal results for Random Forests classification models, as well as competitive results for other classification models. 

The FCQ method was first implemented in the automated machine learning platform using Scala Spark and some key learnings from the implementation are discussed. The feature selection method and its implementation on the platform was proven to be successful for making the model training and prediction more scalable, the feature data pipeline easier to manage and maintain, and the model interpretation more straightforward. The results from one online experiment powered by this platform are provided, demonstrating the business impact. 

It is worth noting that the two extension frameworks proposed in this paper can be further generalized to different variants. For example, for the non-linear association extension, instead of RDC method, other non-linear association measures can be explored, such as the coefficient of determination for nonlinear regression. While for the model-based relevance measure, feature importance scores provided by other models (e.g. Gradient Boosting trees \cite{friedman2002stochastic}) can be used. 

\section*{Acknowledgment}
We would like to extend our gratitude to Fran Bell and Neha Gupta for providing insightful feedback on this paper. We are grateful of the collaboration on use cases and empirical evaluation by Maggie Dou, Yuchen Luo, and Mert Bay. In particular, we appreciate Junjun Li for the guidance and support on implementation. We would also like to thank a number of anonymous reviewers for helpful comments on earlier versions of this paper.

\bibliographystyle{./IEEEtran}
\bibliography{./reference.bib}
\end{document}